\documentclass[11pt,a4paper]{article}
\usepackage[hyperref]{naaclhlt2018}
\usepackage{times}
\usepackage{latexsym}
\usepackage{amsmath,amssymb}
\usepackage{booktabs}
\usepackage{color,soul}
\usepackage{color}
\usepackage{diagbox}
\usepackage{enumitem}
\usepackage{graphicx}
\usepackage{multicol}
\usepackage{multirow}
\usepackage{pifont}
\usepackage{subcaption}
\usepackage{url}

\usepackage{algorithm}
\usepackage{algorithmicx}
\usepackage{algpseudocode}
\usepackage{float}

\usepackage[normalem]{ulem}

\usepackage{color}

\DeclareCaptionFont{10pt}{\fontsize{10pt}{12pt}\selectfont}
\aclfinalcopy 

\title{Incremental Decoding and Training Methods for Simultaneous Translation in Neural Machine Translation}

\author{Fahim Dalvi\thanks{These authors contributed equally to this work} \\
   {\tt faimaduddin@qf.org.qa}
\\ \\
\hspace{72mm} \textbf{Hassan Sajjad \hspace{33mm} Stephan Vogel} \\
\hspace{73mm} {\tt hsajjad@qf.org.qa} \hspace{16mm}  {\tt {svogel}@qf.org.qa}
\\ \\
\hspace{74mm}  Qatar Computing Research Institute -- HBKU \\
  \And
Nadir Durrani\footnotemark[1] \\
 {\tt ndurrani@qf.org.qa} \\
} 

\begin{document}
\maketitle
\begin{abstract}



We address the problem of simultaneous translation by modifying the Neural MT decoder to operate with dynamically built encoder and attention. We propose a tunable agent which decides the best segmentation strategy for a user-defined BLEU loss and Average Proportion (AP) constraint. Our agent outperforms previously proposed Wait-if-diff and Wait-if-worse agents \cite{ChoE16} on BLEU with a lower latency. Secondly we proposed data-driven changes to Neural MT training to better match the incremental decoding framework. 

\end{abstract}

\section{Introduction}

Simultaneous translation is a desirable attribute in 
Spoken Language Translation, 
where 
the translator is required to keep up with the speaker. 
In a lecture or meeting translation scenario where utterances are long, or the end of sentence is not clearly marked, the system must operate on a buffered sequence. Generating translations for such incomplete sequences presents a considerable challenge for machine translation, 
more so in the case of syntactically divergent language pairs (such as German-English), where the context required to correctly translate a sentence, appears much later in the sequence, 
and prematurely committing to a translation 
leads to significant loss in quality. 

Various strategies to select appropriate segmentation points in a streaming input have been proposed 
\cite{FugenWK07, bangalore-EtAl:2012:NAACL-HLT, rangarajansridhar-EtAl:2013:NAACL-HLT,yarmohammadi-EtAl:2013:IJCNLP,oda-EtAl:2014:P14-2}. 
A downside of this approach is that the MT system translates sequences independent of each other, ignoring the context. 
Even if the segmenter decides perfect points to segment the input stream, an MT system requires lexical history to make the correct decision. 


The end-to-end nature of the Neural MT architecture \cite{sutskever2014sequence,bahdanau:ICLR:2015} provides a natural mechanism\footnote{as opposed to the traditional phrase-based decoder (Moses), which requires pre-computation of phrase-table, future-cost estimation \cite{durrani-EtAl:2013:NAACL} and separately maintaining each state-full feature (language model, OSM \cite{durraniEtAl:MT-Summit2015} etc.)}  to integrate stream decoding. Specifically, the recurrent property of the encoder and decoder components provide an easy way to maintain historic context in a fixed size vector. 

\begin{figure*}[!ht]
\centering
	\includegraphics[width=0.9\textwidth]{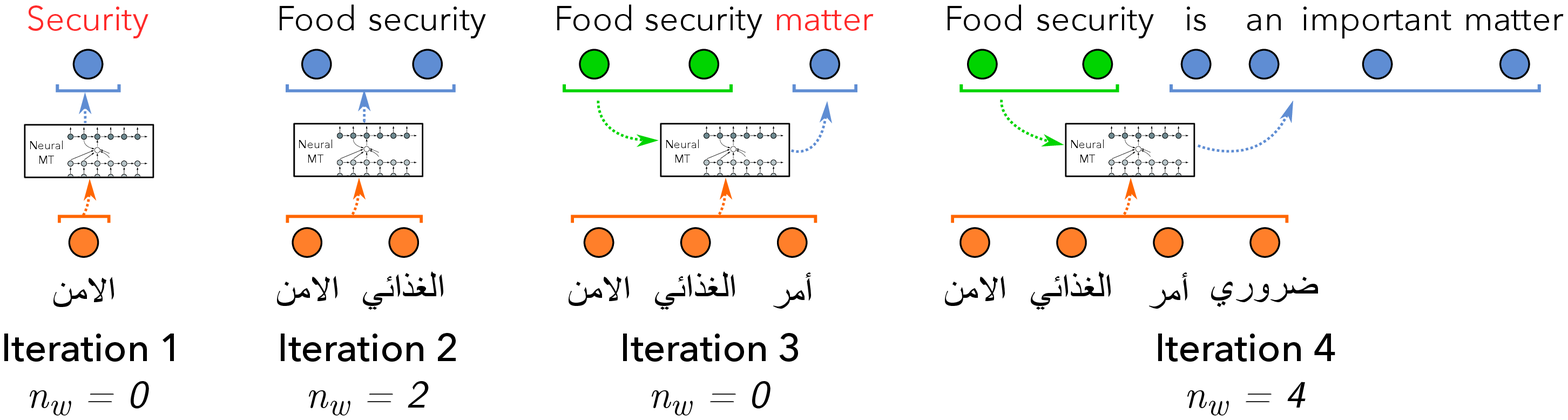}
	\caption{
    A decoding pass over a 4-word source sentence. $n_w$ denotes the number of words the agent chose to commit. Green nodes 
    = committed words, Blue nodes 
    = 
    newly generated words in the current iteration. Words marked in red 
    are discarded, as the agent chooses to not commit them. 
    }
	\label{fig:stream-iterations}
\end{figure*}

We modify the neural MT architecture to operate in an online fashion where i) the \emph{encoder} and the \emph{attention} are updated dynamically as new input words are added, through a {\tt READ} operation, and ii) the \emph{decoder} generates 
output from the available encoder states, through a {\tt WRITE} operation. The decision of when to {\tt WRITE} is learned through a tunable segmentation agent, based on user-defined thresholds. 
Our incremental decoder 
significantly outperforms 
the chunk-based 
decoder and restores the oracle performance with a deficit of $\le 2$ BLEU points across 4 language pairs with a moderate delay. We additionally explore whether 
modifying the Neural MT training to match the decoder can improve performance. 
While we observed significant restoration in the case of chunk decoding matched with chunk-based NMT training, the same was not found true with our proposed incremental training to match the incremental decoding framework. 


The remaining paper is organized as follow: Section \ref{sec:IncDecoding} describes 
modifications to  the NMT decoder to enable stream decoding. Section \ref{sec:strategies} describes various agents to learn a {\tt READ/WRITE} strategy. Section \ref{sec:results} presents evaluation and results. Section \ref{sec:IncTraining} describes modifications to the NMT training to mimic corresponding decoding strategy,  
and Section \ref{sec:conclusion} concludes the paper.




\section{Incremental Decoding}
\label{sec:IncDecoding}

\paragraph{Problem:}

In a stream decoding scenario, the entire source sequence is not readily available. The translator must either wait for the 
sequence to finish in order to compute the encoder state, or commit partial translations at several intermediate steps, potentially losing contextual information. 


\paragraph{Chunk-based Decoder:} A straight forward way to enable simultaneous translation is to chop the incoming input after every N-tokens. 
A drawback of 
these approaches is that the translation and segmentation process operate independently of each other, and the previous contextual history 
is not considered when translating the current chunk. This information is important to generate grammatically correct and coherent translations. 

\paragraph{Incremental Decoding:} The RNN-based 
NMT framework provides a natural mechanism to preserve context 
and accommodate streaming. 
The 
decoder 
maintains the entire target 
history through the previous decoder state alone. 
But to enable incremental neural decoding, we have to address the following constraints: i) 
how 
to dynamically build the encoder and attention with the streaming input? 
ii) 
what is the best strategy to pre-commit translations at several intermediate points?


\label{sec:IncDecoder}
\begin{algorithm}[t]
\caption{Algorithm for incremental decoder}
\label{alg:decoder}
\small
\begin{algorithmic}
	\State $s$, Source sequence
	\State $s'$, Available source sequence
    \State $t_c$, Committed target sequence
    \State $t$, Current decoded sequence for $s'$
    \State $n_w$, Number of tokens to commit
    \Statex 
    \State $s' \gets $ empty
	\For{$token$ in $s$}
    	\Comment{{\tt READ} operation}
    	\State $s' \gets s' + token$
        \State $t \gets $\Call{NMTDecoder}{$s'$, $t_c$}
        \If {$s' \neq s$}
        	\State $n_w \gets$\Call{Agent}{$s'$, $t_c$, $t$}
        \Else
        	\State $n_w \gets$ length$(t) - $ length$(t_c)$
        \EndIf
        \Comment{commit all new words if we have seen the entire source}
        \State $t_c' \gets$\Call{GetNewTokens}{$t_c$, $t$, $n_w$}
        \State $t_c \gets t_c + t_c'$
        \Comment{{\tt WRITE} operation}
    \EndFor
    \Statex
    \Function{GetNewTokens}{$t_c$, $t$, $n_w$}
    	\State $start \gets $length$(t_c)+1$
        \State $end \gets start + n_w$
    	\State \Return $t[start : end]$
    \EndFunction
\end{algorithmic}
\end{algorithm}


\noindent Inspired by 
\newcite{ChoE16}, we modify the NMT decoder to operate in a sequence of {\tt READ} and {\tt WRITE} operations. 
The 
former reads the next word from the buffered source sequence and translates it 
using the available 
context, 
and the latter is computed through an {\normalsize{A}\scriptsize{GENT}}, which decides how many 
words should be committed from this generated translation. Note that, when a translation is generated in the {\tt READ} operation, the already committed target words remain unchanged, i.e. the generation is continued from the last committed target word 
using the saved decoder state. See Algorithm \ref{alg:decoder} for details.
The {\normalsize{A}\scriptsize{GENT}} 
decides how many target words to 
{\tt WRITE} after every {\tt READ} operation, 
and has complete control over the context each target word gets to see before being committed,
as well as 
the overall delay 
incurred.  Figure \ref{fig:stream-iterations} shows the incremental decoder in action, where the agent decides to not commit any target words in iterations 1 and 3. The example 
shows an 
instance where the incorrectly translated words are discarded when more 
context becomes available. Given this generic framework, 
we describe several {\normalsize{A}\scriptsize{GENTS}} in Section \ref{sec:strategies}, 
trained to optimize the BLEU loss and latency. 

 	


\begin{figure}[!t]
	\centering
	\includegraphics[width=0.8\linewidth]{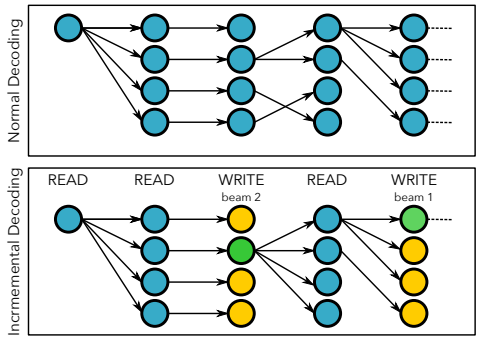}
	\caption{Beam Search in normal decoding vs incremental decoding. Green nodes indicate the hypothesis selected by the agent to \texttt{WRITE}. Since we cannot change what we have already committed, the other nodes (marked in yellow) are discarded and future hypotheses originate from the selected hypothesis alone. Normal beam search is executed for \emph{consecutive} \texttt{READ} operations (blue nodes).}
	\label{fig:stream-beam}
\end{figure}

\paragraph{Beam Search:}

Independent of the agent being used, the modified NMT architecture incurs some complexities for beam decoding. For 
example, if at some iteration the decoder generates $5$ new words, but the agent decides to commit only $2$ of these, the best hypothesis at the $2^{nd}$ word may not be the same as the one at the $5^{th}$ word. Hence, the agent 
has to re-rank the hypotheses at the last target word it decides to commit. Future 
hypotheses then continue from this selected hypothesis. 
See Figure \ref{fig:stream-beam} for a visual representation. 
The overall utility of beam decoding is reduced in the case of incremental decoding, because it is necessary to commit and retain only one beam at several points to start producing output with minimal delay.

\section{Segmentation Strategies}
\label{sec:strategies}
\begin{figure*}[!ht]
    \centering
    \begin{subfigure}[b]{0.49\textwidth}
        \includegraphics[width=\textwidth]{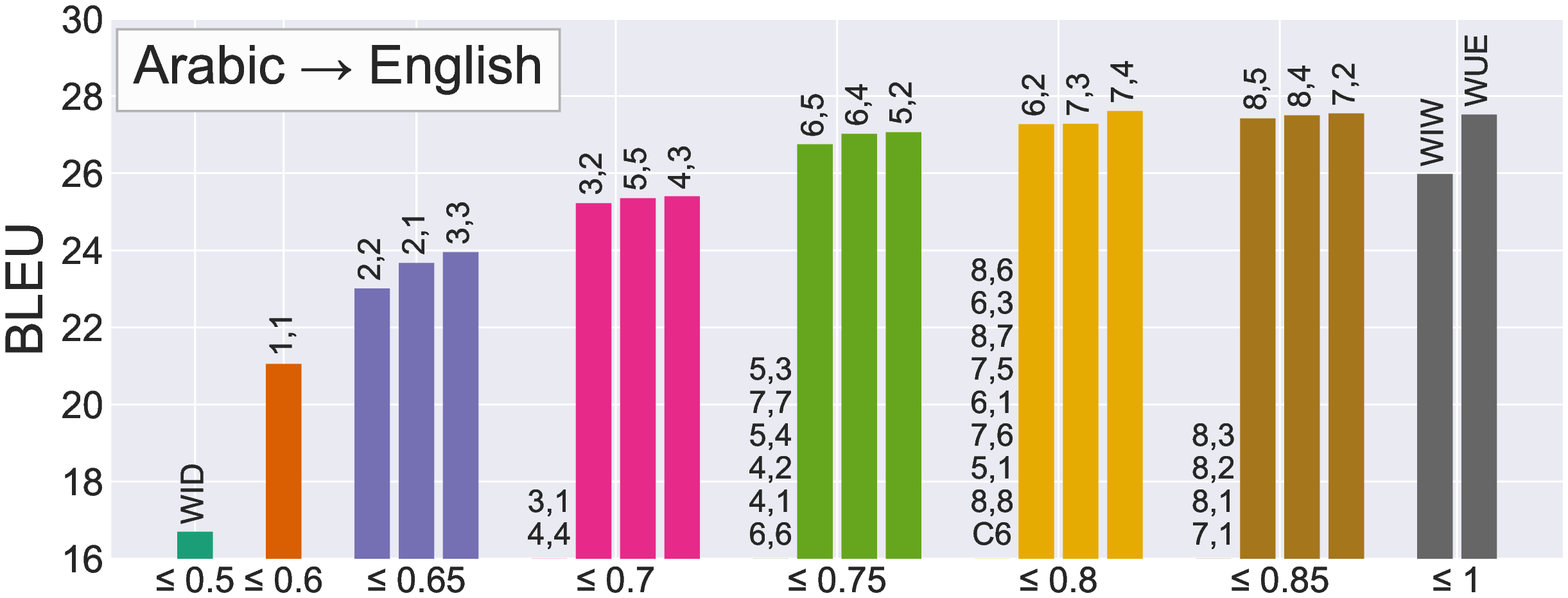}
        \label{fig:arabic-grid}
    \end{subfigure}
    \hfill
    \begin{subfigure}[b]{0.49\textwidth}
        \includegraphics[width=\textwidth]{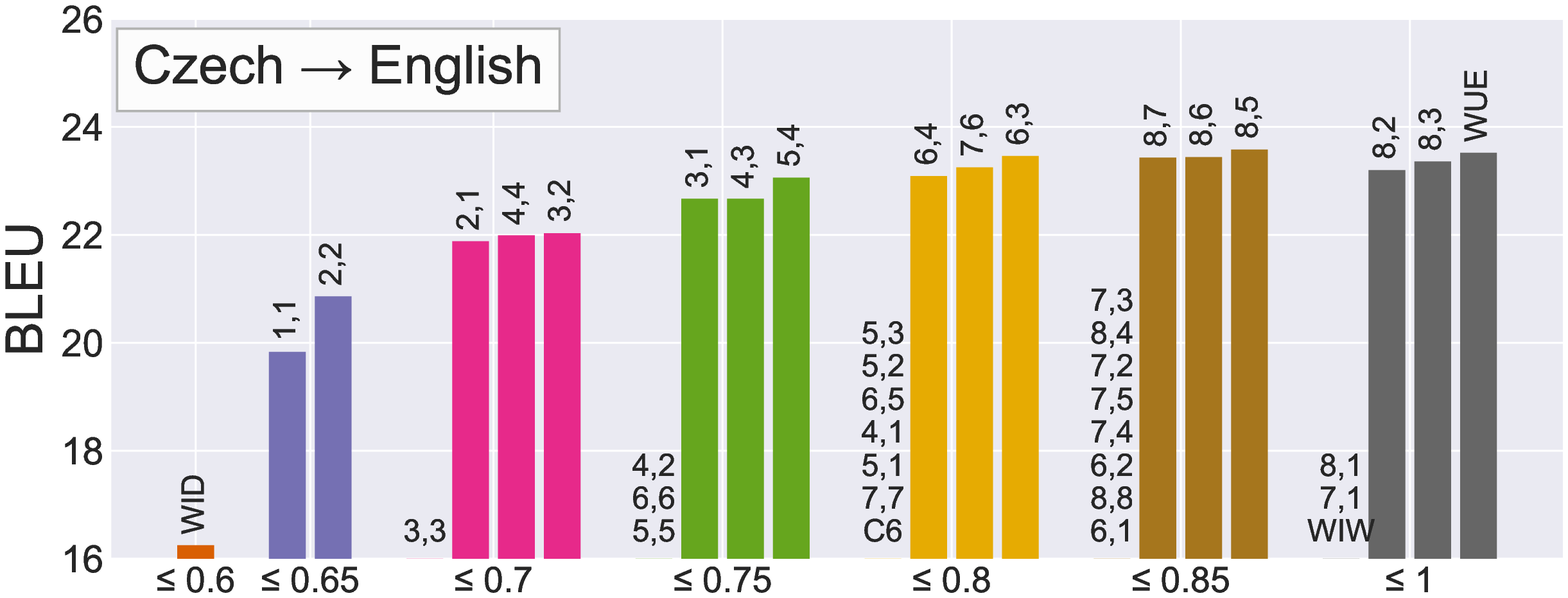}
        \label{fig:czech-grid}
    \end{subfigure}\\[-1.5em]
    \begin{subfigure}[b]{0.49\textwidth}
        \includegraphics[width=\textwidth]{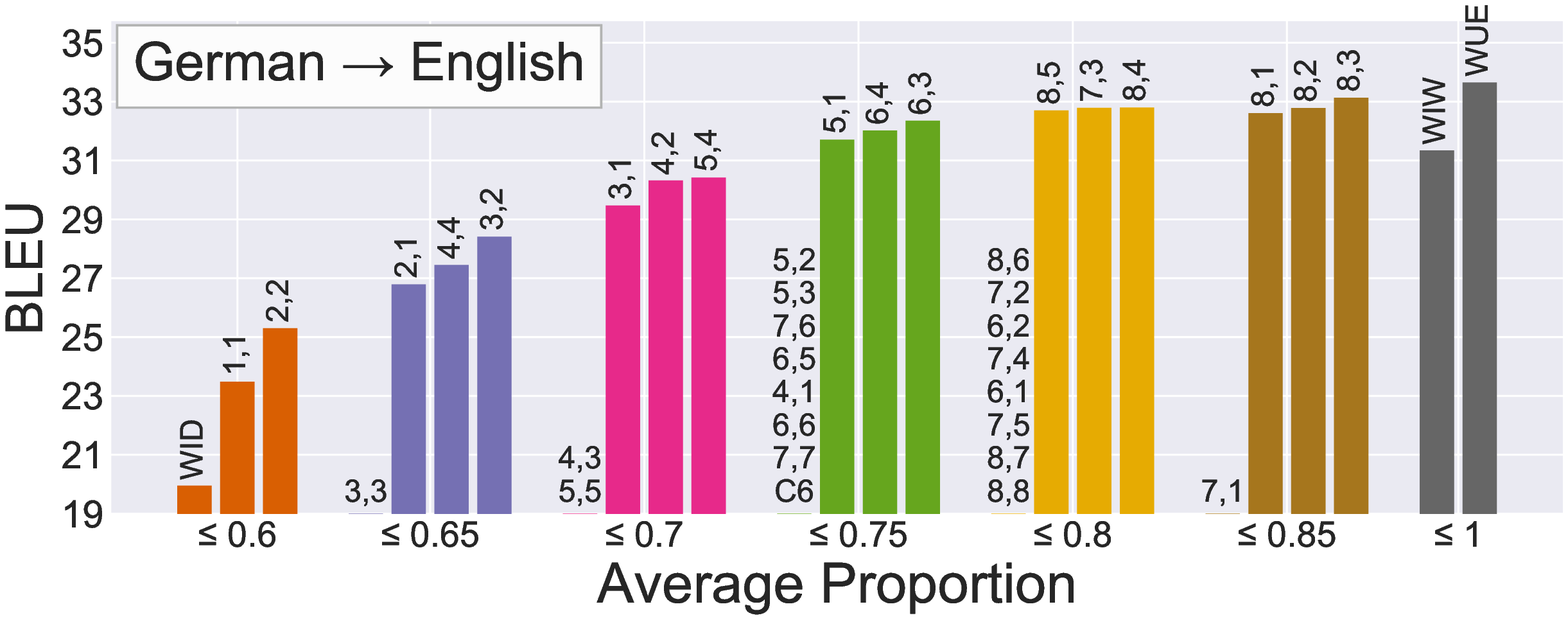}
        \label{fig:german-grid}
    \end{subfigure}
    \hfill
    \begin{subfigure}[b]{0.49\textwidth}
        \includegraphics[width=\textwidth]{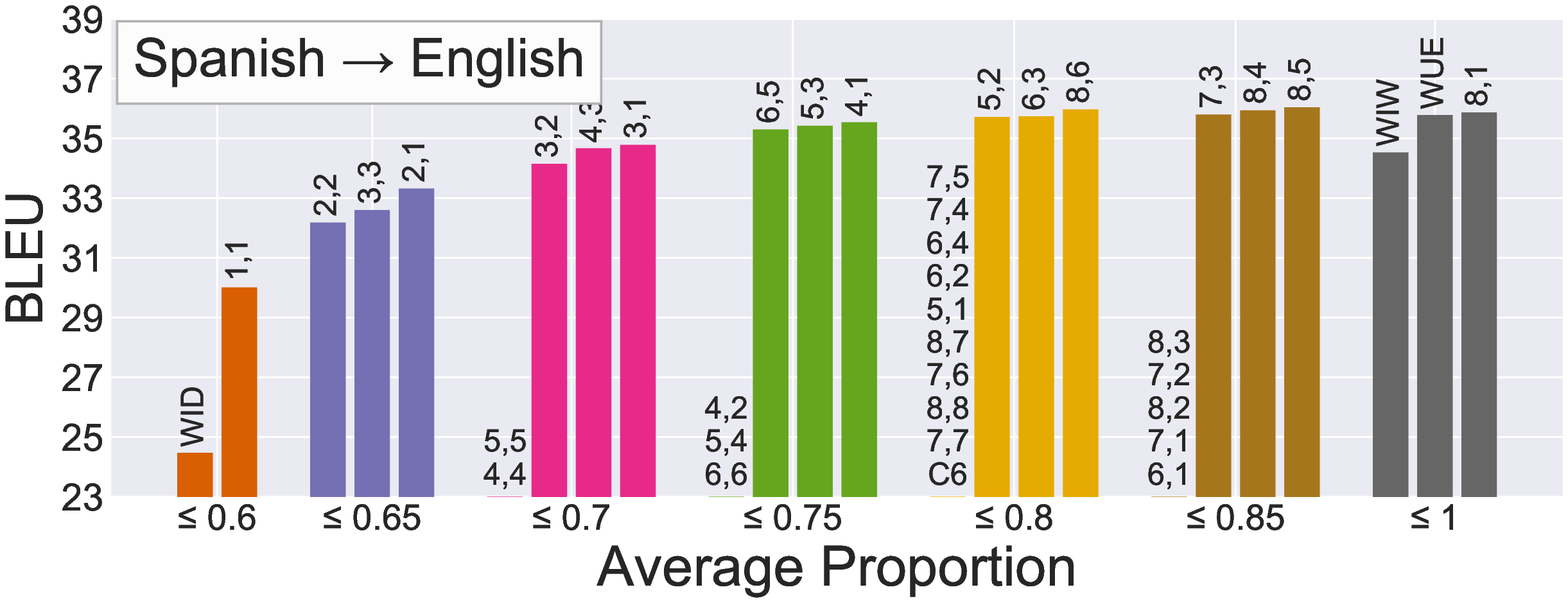}
        \label{fig:spanish-grid}
    \end{subfigure}
    \caption{Results for various streaming {\normalsize{A}\scriptsize{GENTS}} ({\tt WID}, {\tt WIW}, {\tt WUE}, {\tt C6} (Chunk decoding with a N=6) and $\mathcal{S}$,$\mathcal{RW}$ for {\tt STATIC-RW}) on the tune-set. 
    For each AP bucket, we only show the Agents with the top 3 BLEU scores in that bucket, with 
remaining listed in descending order of their BLEU scores.}\label{fig:ID-grid}
\end{figure*}

In this section, 
we discuss different {\normalsize{A}\scriptsize{GENTS}} that we evaluated in our modified incremental decoder. To measure latency in these agents, we use \emph{Average Proportion} (AP) metric as defined by \newcite{ChoE16}. AP is calculated as the total number of source words each target word required before being committed, normalized by the product of the source and target lengths. It varies between $0$ and $1$ with lesser being better. 
See supplementary material for details. 

\paragraph{Wait-until-end:}
The {\tt WUE} agent waits for the entire source sentence before decoding, and serves as an upper bound on the performance of our agents, albeit with the worst $AP=1$. 


\paragraph{Wait-if-worse/diff:}
We reimplemented the baseline 
agents described in \newcite{ChoE16}. The \textbf{Wait-if-Worse} ({\tt WIW}) agent {\tt WRITES} a target word if its probability does not decrease after a {\tt READ} operation. The \textbf{Wait-if-Diff} ({\tt WID}) agent instead {\tt WRITES} a target word if the target word remains unchanged after a {\tt READ} operation. 

\paragraph{Static Read and Write:}
The {\tt STATIC-RW:} agent is inspired from the chunk-based decoder and tries to resolve its shortcomings while maintaining its simplicity. The primary drawback of the chunk-based decoder is the loss of context across chunks. Our agent starts by performing $\mathcal{S}$ {\tt READ} operations, followed by repeated $\mathcal{RW}$ {\tt WRITES} and {\tt READS} until the end of the source sequence. The number of {\tt WRITE} and {\tt READ} operations is the same to ensure that the gap between the source and target sequence does not increase with time. The initial $\mathcal{S}$ {\tt READ} operations essentially create a buffer of $\mathcal{S}$ tokens, allowing some future context to be used by the decoder. Note that the latency induced by this agent in this case is only in the beginning, and remains constant for the rest of the sentence.
This method actually introduces a class of {\normalsize{A}\scriptsize{GENTS}} based on their $\mathcal{S}$,$\mathcal{RW}$ values. We tune $\mathcal{S}$ and $\mathcal{RW}$ to select the specific {\normalsize{A}\scriptsize{GENT}} with the user-defined BLEU-loss and AP thresholds.  

\section{Evaluation}
\label{sec:results}

\paragraph{Data:} We trained systems for 
4 language pairs: 
German-, Arabic-, Czech- and Spanish-English pairs using the data 
made available for 
IWSLT \cite{cettolo2014report}. See supplementary material for data stats. These language pairs present a diverse set of challenges for this problem, with Arabic and Czech being morphologically rich, German being syntactically divergent, and Spanish introducing local reorderings
with respect to English.


\paragraph{NMT System:} We trained a 2-layered LSTM encoder-decoder models with attention using the \texttt{seq2seq-attn} implementation. Please see supplementary material for settings. 

\begin{figure}
	\centering
	\includegraphics[width=0.96\linewidth]{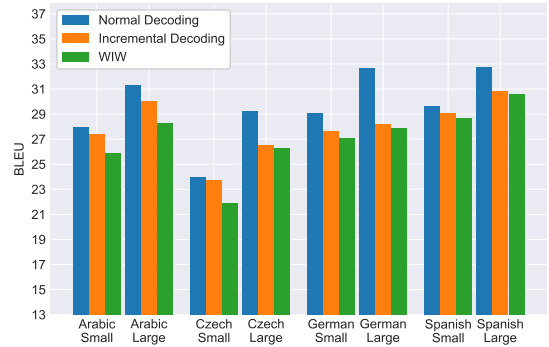}
	\caption{Averaged results on test-sets (2011-2014) using the models trained on small and large datasets using $AP \le 0.75$. Detailed test-wise results are available in the supplementary material.}
	\label{fig:scaled}
\end{figure}

\paragraph{Results:} Figure \ref{fig:ID-grid} shows the results of 
various streaming agents. Our proposed {\tt STATIC-RW} agent outperforms other methods while maintaining an AP $<0.75$
with a loss of less than $0.5$ BLEU points on Arabic, Czech and Spanish. This was found to be consistent for all test-sets 2011-2014 (See under ``small" models in Figure \ref{fig:scaled}). In the case of German the loss at AP $<0.75$ was around 
$1.5$ BLEU points. The syntactical divergence and rich morphology of German posits a bigger challenge and requires larger context than other language pairs. For example the conjugated verb in a German verb complex appears in the second position, while the main verb almost always occurs at the end of the sentence/phrase \cite{durrani-EtAl:2013:WMT2}. Our methods are also comparable to the more sophisticated techniques involving Reinforcement Learning to learn an agent introduced by \newcite{gu-EtAl:2017:EACLlong} and \newcite{Satija-Pineau}, but without the overhead of expensive training for the agent.


\paragraph{Scalability:} \label{sec:scale} 
The preliminary results were obtained using models trained on the 
TED corpus only. We conducted further experiments by training models on larger data-sets (See the supplementary section again for data sizes) to see if our findings are scalable. 
We 
fine-tuned \cite{luong-manning:iwslt15,sajjad-etal:iwslt17} our models with the in-domain data to avoid domain disparity. We then re-ran our agents with the best $\mathcal{S}$,$\mathcal{RW}$ values (with an AP under $0.75$) for each language pair. 
Figure \ref{fig:scaled} (``large" models) shows that 
the BLEU loss from the respective oracle increased 
when the models were trained with bigger data sizes. This could be attributed to the increased lexical ambiguity from the large amount of out-domain data, which can only be resolved with additional contextual information. However our results were still better than the WIW agent, which also has an AP value above $0.8$. 
Allowing similar AP, 
our {\tt STATIC-RW} agents were able to restore the BLEU loss to be $\leq$ 1.5 for all language pairs except German-English. Detailed test results are available in the suplementary material.


\section{Incremental Training}
\label{sec:IncTraining}

The limitation of previously described decoding approaches (chunk-based and incremental) is the mismatch between training and decoding. 
The training is carried on full sentences, however, at the test time, the decoder generates hypothesis 
based on incomplete source information.
This discrepancy between training and decoding 
can be potentially harmful. In Section \ref{sec:IncDecoder}, we presented two methods to address the partial input sentence decoding problem, the \emph{Chunk Decoder} and the \emph{Incremental Decoder}. We now train models to match the corresponding decoding scenario. 

\subsection{Chunk Training}

In chunk-based training, we simply split each training sentence into chunks of $N$ tokens.\footnote{Although randomly segmenting the source sentence based on number of tokens is a na\"ive approach that does not take into account the linguistic properties, our goal here was to exactly match the training with the chunk-based decoding scenario.
} The corresponding target sentence for each chunk is generated by having a span of target words that are word-aligned\footnote{We used fast-align \cite{dyer-chahuneau-smith:2013:NAACL-HLT} for  alignments.} with the words in the source span. 
Chunking the data into smaller segments increases the training time significantly. To overcome this problem, we train a model on the full sentences using all the data and then fine-tune it with the in-domain chunked data.

\subsection{Add-M Training}

Next we formulate a training mechanism to match the incremental decoding described in Section \ref{sec:IncDecoder}. A way to achieve this is to force the attention on a local span of encoder states and block it from giving weight to the non-local (rightward) encoder states. The hope is that in the case of long-range dependencies, the model 
learns to predict these dependencies without the entire source context. 
Such a training procedure is non-trivial, as it requires 
dynamic inputs to the attention mechanism while training, including backpropagation where some encoder states which have been seen by the attention mechanism a greater number of times dynamically receiving more gradient inputs. We leave this idea as future work, while 
focusing on a data-driven technique to mimic this kind of training as described below. 

We start with the first $N$ words in a source sentence and generate target words that are aligned to these words.
We then generate the next training instances with $N+M$, $N+2M$, $N+3M$ ... source words until the end of sentence has been reached.\footnote{We trained with $N=6$ and $M=1$ for our experiments.}
The resulting training roughly mimics the decoding scenario where the source-side context is gradually built. 
The down-side of this method is that the data size increases quadratically,
 making the training infeasible. To overcome this, we fine-tune a model trained on full sentences with the in-domain corpus generated using this method.

\begin{figure}
	\centering
	\includegraphics[width=0.96\linewidth]{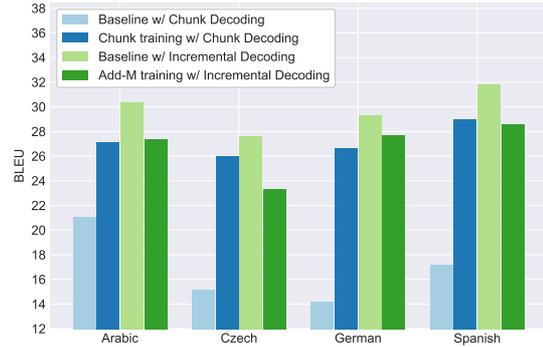}
	\caption{Averaged test set results on various training modifications}
	\label{fig:training}
\end{figure}

\subsection{Results}

The results in Figure \ref{fig:training} show that matching the chunk-decoding with corresponding chunk-based training significantly improves performance, with a gain of up to 12 BLEU points. 
However, we were not able to improve upon our incremental decoder, with the results deteriorating notably. One reason for this degradation is that the training/decoding scenarios are still not perfectly matched. The training pipeline in this case also sees the beginning of sentences much more often, which could lead to unnatural distributions being inferred within the model.

\section{Conclusion}
\label{sec:conclusion}
We addressed the problem of simultaneous translation by modifying the architecture in Neural MT decoder. We presented a tunable agent which decides the best segmentation strategy based on user-defined BLEU loss and AP constraints. Our results showed improvements over previously established WIW and WID methods. We additionally 
modified the Neural MT training to match the incremental decoding, which significantly improved the chunk-based decoding, but we did not observe any improvement using \emph{Add-M Training}. The code for our incremental decoder and agents has been made available.\footnote{\url{https://github.com/fdalvi/seq2seq-attn-stream}}
While were able to significantly improve the the chunk-based decoder, we did not observe any improvement using the \emph{Add-M Training}. In the future we would like to change the training model to dynamically build the encoder and the attention model in order to match our incremental decoder.

\bibliography{naaclhlt2018}
\bibliographystyle{apalike}

\cleardoublepage
\appendix

\section{Supplementary Material}
\label{sec:supplemental}

\subsection{Data and Preprocessing}

We trained systems for four language pairs namely German-English, Arabic-English, Czech-English and Spanish-English using the data for the translation task of the International Workshop on Spoken Language Translation 
\cite{cettolo2014report}. Apart from using the in-domain TED corpus ($\approx$ 200K sentences), we additionally used Europarl and News Corpus made available for the recent WMT campaign. For Arabic-to-English, we also  news Corpus and a subset of UN corpus (1 Million sentences) \cite{Un_2010_lrec}. We used a concatenation of dev- and test-2010 for tuning Neural MT models, test-2011 for development (tuning the Static Read and Write agent) and tests 2012-14 for testing. We used Moses \cite{Moses:2007} preprocessing pipeline including tokenization and truecasing. For Arabic we used Farasa segmentation \cite{abdelali-EtAl:2016:N16-3} with BPE \cite{sennrich-haddow-birch:2016:P16-12} as suggested in \cite{sajjad:2017:ACL}. We trained the BPE models separately for both 
the source and target datasets instead of jointly training limiting the number of operations to 49,500, as suggested in \cite{sennrich-haddow-birch:2016:P16-12}.

\begin{table}[h]
\small
\centering
\resizebox{\linewidth}{!}{
\begin{tabular}{l|r|r|rrrr}
\toprule
Pair & ID & Cat & test11 & test12 & test13 & test14 \\ 
\midrule
ar-en & 229K & 1.26M  & 1199 & 1702 &  1169 & 1107 \\
& 4.4M  & 28.7M & 22K & 28K & 24K & 20K  \\
& 4.7M & 30.2M & 26K & 32K & 28K & 24K  \\
\midrule
de-en & 209K  & 2.4M & 1433 & 1700 &  993 & 1305 \\
& 4.0M  & 61.9M & 26K & 29K & 20K & 24K  \\
& 4.3M & 64.7M & 27K & 31K & 21K & 25K  \\
\midrule
cs-en & 122K & 900K  & 1013 & 1385 &  1327 & -- \\
& 2.0M & 20.3M & 15K & 21K & 24K &  -- \\
& 2.5M & 23.6M & 18K & 25K & 28K &  -- \\
\midrule
es-en & 188K & 2.3M & 1435 & 1385 &  -- & -- \\
& 3.6M & 66.9M & 25K & 27.5K & -- &  -- \\
& 3.8M & 64.4M & 27K & 31K & -- & --  \\
\bottomrule
\end{tabular}
}
\caption{\label{tab:data} Data Statistics: First Row = Number of Sentences, Second Row: Number of Tokens in Source Language, Third Row: Number of Tokens in Target Language. First Column = statistics for the in-domain TED corpus, Second Column = Statistics for the Concatenated Data}. 
\end{table}


\subsection{Neural MT system} 

We train a 2-layer LSTM encoder-decoder with attention using the \texttt{seq2seq-attn} implementation with the following settings: word vectors and LSTM states with 500 dimensions, SGD with an initial learning rate of 1.0, a decay rate of 0.5, and dropout rate of 0.3. The MT systems are trained for 13 epochs. We used uni-directional encoder because it is not possible to compute the encoder in right-to-left direction in the streaming scenario, due to unavailability of the full input sentence. Computing right-to-left encoder states with whatever input sequence is available is also not viable as it requires expensive re-computation after each input word is added.\footnote{Unlike left-to-right encoder which only requires single computation after each input word is added.} We also trained the models by initializing the first decoder state with zeros, rather than using the final encoder state, which will not be available during stream decoding.


\subsection{Average Proportion}

In normal decoding, the BLEU metric is commonly used to calculate the quality of translations from a system. In stream decoding, we have to also consider the delay induced by the system along with its BLEU. In our work, we use \emph{Average Proportion} (AP) as defined by \newcite{gu-EtAl:2017:EACLlong}. AP is calculated as the total number of source words each target word required before being committed, normalized by the product of the source and target lengths. Formally, if $s(t_i)$ is the number of source words required for target word $i$ before being committed, $X$ is the source sequence and $Y$ is the generated target sequence:

\begin{align}
AP &= \frac{1}{|X| \cdot |Y|} \sum_{t_i}^{Y} s(t_i)
\end{align}

\subsection{Incremental Decoder}
Figure \ref{fig:scaled} shows the average results on the test-sets for the models trained on in-domain TED corpus. Here, we present the test-wise results for the interested reader. Missing table values correspond to unavailable test-sets on the IWSLT webpage. See Table \ref{tab:incDec}.

\begin{table*}[ht!]
  \small
  \centering
  	\begin{tabular}{l|r|ccc||r|ccc}
      \toprule
      Pair & Agent & test12 & test13 & test14 & Agent & test12 & test13 & test14\\ 
      \midrule
      ar-en & {\tt WUE} & 30.16 & 28.16 & 25.53 & {\tt WUE} & 32.84 & 32.23 & 28.95  \\
			& $5,2$     & 29.31 & 27.72 & 25.21 & $7,2$ & 31.71 & 31.46 & 28.29  \\
			& {\tt WIW} & 28.06 & 25.86 & 23.75 & {\tt WIW} & 29.48 & 28.82 & 26.52 \\
			& {\tt WID} & 19.89 & 17.24 & 15.64 &  & & &  \\
      \midrule
      cs-en & {\tt WUE} & 22.95 & 25.03 & -- & {\tt WUE} & 27.97 & 30.50 & -- \\
            & $5,4$     & 22.97 & 24.46 & -- & $8,3$ & 26.68 & 29.37 & -- \\
            & {\tt WIW} & 21.78 & 21.99 & -- & {\tt WIW} & 25.20 & 27.43 & -- \\
            & {\tt WID} & 16.37 & 17.07 & -- & & & &  \\
      \midrule
      de-en & {\tt WUE} & 29.20 & 31.31 & 26.61 & {\tt WUE} & 35.52 & 35.01 & 30.44 \\
			& $6,3$     & 27.94 & 29.90 & 25.07 & $8,3$ & 28.62 & 31.71 & 27.09 \\
			& {\tt WIW} & 27.77 & 29.55 & 23.88 & {\tt WIW} & 27.94 & 30.05 & 25.56 \\
			& {\tt WID} & 19.15 & 20.73 & 16.46 & & & & \\
      \midrule
      es-en & {\tt WUE} & 29.65 & -- & -- & {\tt WUE} & 32.78 & -- & --  \\
            & $4,1$     & 29.04 & -- & -- & $8,1$ & 32.05 & -- & -- \\
            & {\tt WIW} & 28.65 & -- & -- & {\tt WIW} & 30.59 & -- & -- \\
            & {\tt WID} & 21.90 & -- & -- & & & & \\
      \bottomrule
    \end{tabular}
\caption{\label{tab:incDec} Left Side: Test-wise results for "Small" models in Figure 4, Right Side: Test-wise results for "Large" models in Figure 4}. 
\end{table*}

\subsection{Scalability}

In section \ref{sec:scale} we note that even though the {\tt WIW} agent's performance is not significantly below our selected {\tt STATIC-RW} agent, its AP is much higher. When we allow our {\tt STATIC-RW} agent an AP similar to that of the {\tt WIW} agent, we are able to restore the BLEU loss to be less than 1.5 for all language pairs except German-English. Here are the results in detail. See Table \ref{tab:incDec}.

\end{document}